\documentclass[conference]{IEEEtran}
\usepackage{times}
\usepackage{graphicx}
\usepackage{color}
\usepackage{float}
\usepackage{caption}
\usepackage{xcolor}
\usepackage{colortbl}
\usepackage{subcaption}
\usepackage{amsmath}
\usepackage{amssymb} 
\usepackage{pifont}
\usepackage{multirow}
\usepackage{tcolorbox}
\usepackage{listings}
\usepackage{makecell}
\usepackage{pifont}
\usepackage{booktabs}
\usepackage{fancyhdr}

\usepackage[numbers]{natbib}
\usepackage{multicol}
\usepackage[bookmarks=true]{hyperref}

\newcommand{\numembodiments}{15\ }
\newcommand{\numtasks}{421\ }
\newcommand{\numdemonstrations}{180,000\ }
\newcommand{\numdemonstrationsabs}{180K+\ }
\newcommand{\numscenarios}{16\ }
\newcommand{\numactions}{39\ }
\newcommand{\numobjects}{432\ }

\definecolor{lightblue}{RGB}{0, 153, 255}
\definecolor{lightgreen}{RGB}{0, 200, 102}
\definecolor{lightred}{RGB}{255, 52, 52}

\lstdefinestyle{yaml}{
     basicstyle=\color{blue}\footnotesize,
     rulecolor=\color{black},
     string=[s]{'}{'},
     stringstyle=\color{blue},
     comment=[l]{:},
     commentstyle=\color{black},
     morecomment=[l]{-}
}

\makeatletter
\DeclareRobustCommand*{\IEEEauthorrefmark}[1]{%
  \raisebox{0pt}[0pt][0pt]{\textsuperscript{\footnotesize #1}}%
}
\makeatother

\begin{document}
\title{\textcolor{orange}{Robo}\textcolor{lightblue}{CO}\textcolor{lightgreen}{IN}: An Open-Sourced Bimanual \textcolor{orange}{Robo}tic \\ Data \textcolor{lightblue}{CO}llection for \textcolor{lightgreen}{IN}tegrated Manipulation}

\author{
\authorblockN{
  Shihan Wu\authorrefmark{1,2*\dag},\,
  Xuecheng Liu\authorrefmark{1*\dag},\,
  Shaoxuan Xie\authorrefmark{1*\dag},\,
  Pengwei Wang\authorrefmark{1*},\,
  Xinghang Li\authorrefmark{1*},\, \\
  Bowen Yang\authorrefmark{3},\, 
  Zhe Li\authorrefmark{3},\,
  Kai Zhu\authorrefmark{3},\,
  Hongyu Wu\authorrefmark{1,4},\,
  Yiheng Liu\authorrefmark{1,4},\,
  Zhaoye Long\authorrefmark{1,5},\, 
  Runtian Xu\authorrefmark{1,4},\, \\ 
  Yue Wang\authorrefmark{1},\,
  Chong Liu\authorrefmark{1,4},\, 
  Dihan Wang\authorrefmark{1,4},\, 
  Ziqiang Ni\authorrefmark{1},\,
  Xiang Yang\authorrefmark{1},\,
  You Liu\authorrefmark{1},\,
  Ruoxuan Feng\authorrefmark{1,6},\, \\
  Lei Zhang\authorrefmark{1,7},\, 
  Denghang Huang\authorrefmark{1,8},\, 
  Chenghao Jin\authorrefmark{1,9},\,
  Anlan Yin\authorrefmark{1,10},\,
  Xinlong Wang\authorrefmark{1},\, 
  Zhenguo Sun\authorrefmark{1},\, \\
  Junkai Zhao\authorrefmark{1},\, 
  Mengfei Du\authorrefmark{1},\, 
  Mingyu Cao\authorrefmark{1},\,
  Xiansheng Chen\authorrefmark{1},\,
  Hongyang Cheng\authorrefmark{1},\,
  Xiaojie Zhang\authorrefmark{1},\, \\ 
  Yankai Fu\authorrefmark{1,11},\, 
  Ning Chen\authorrefmark{1,11},\,
  Cheng Chi\authorrefmark{1},\,
  Sixiang Chen\authorrefmark{1,11},\,
  Huaihai Lyu\authorrefmark{1,7},\,
  Xiaoshuai Hao\authorrefmark{1},\, \\
  Yequan Wang\authorrefmark{1},\, 
  Bo Lei\authorrefmark{1},\, 
  Dong Liu\authorrefmark{1},\, 
  Xi Yang\authorrefmark{1},\,
  Yance Jiao\authorrefmark{1},\,
  Tengfei Pan\authorrefmark{1},\,
  Yunyan Zhang\authorrefmark{1},\, \\
  Songjing Wang\authorrefmark{1},\, 
  Ziqian Zhang\authorrefmark{1},\, 
  Xu Liu\authorrefmark{1},\, 
  Ji Zhang\authorrefmark{12},\, 
  Caowei Meng\authorrefmark{13},\,
  Zhizheng Zhang\authorrefmark{13,1},\, \\
  Jiyang Gao\authorrefmark{14},\, 
  Song Wang\authorrefmark{15},\,
  Xiaokun Leng\authorrefmark{15},\, 
  Zhiqiang Xie\authorrefmark{16},\, 
  Zhenzhen Zhou\authorrefmark{16},\,
  Peng Huang\authorrefmark{17},\, \\
  Wu Yang\authorrefmark{17},\, 
  Yandong Guo\authorrefmark{18},\, 
  Yichao Zhu\authorrefmark{18},\, 
  Suibing Zheng\authorrefmark{19},\, 
  Hao Cheng\authorrefmark{20},\,
  Xinmin Ding\authorrefmark{21},\, \\
  Yang Yue\authorrefmark{22},\, 
  Huanqian Wang\authorrefmark{22},\, 
  Chi Chen\authorrefmark{22},\, 
  Jingrui Pang\authorrefmark{1,22},\, 
  YuXi Qian\authorrefmark{23},\, 
  Haoran Geng\authorrefmark{24},\,
  Lianli Gao\authorrefmark{2},\, \\
  Haiyuan Li\authorrefmark{4},\, 
  Bin Fang\authorrefmark{4,1},\, 
  Gao Huang\authorrefmark{22,1},\, 
  Yaodong Yang\authorrefmark{11,1,25},\, 
  Hao Dong\authorrefmark{11,1},\,
  He Wang\authorrefmark{11,13,1},\, \\
  Hang Zhao\authorrefmark{22,14},\, 
  Yadong Mu\authorrefmark{11,1},\, 
  Di Hu\authorrefmark{6,1},\, 
  Hao Zhao\authorrefmark{22,1},\, 
  Tiejun Huang\authorrefmark{11,1},\, 
  Shanghang Zhang\authorrefmark{11,1\ddag},\, \\
  Yonghua Lin\authorrefmark{1\ddag},\,
  Zhongyuan Wang\authorrefmark{1\ddag},
  Guocai Yao\authorrefmark{1\dag\ddag}
}
\vspace{1mm}
\authorblockA{
  \authorrefmark{1}Beijing Academy of Artificial Intelligence,
  \authorrefmark{2}University of Electrical Science and Technology of China, \\
  \authorrefmark{3}Ant Digital Technologies, Ant Group,
  \authorrefmark{4}Beijing University of Posts and Telecommunications, \\
  \authorrefmark{5}Harbin Institute of Technology,
  \authorrefmark{6}Renmin University of China, 
  \authorrefmark{7}Chinese Academy of Sciences, \\
  \authorrefmark{8}Huazhong University of Science and Technology,
  \authorrefmark{9}University of Cambridge, 
  \authorrefmark{10}Harbin Engineering University, \\
  \authorrefmark{11}Peking University,
  \authorrefmark{12}Southwest Jiaotong University,
  \authorrefmark{13}Galbot,
  \authorrefmark{14}Galaxea,
  \authorrefmark{15}Leju Robotics,
  \authorrefmark{16}Agilex Robotics, \\
  \authorrefmark{17}TQ-Artisan,
  \authorrefmark{18}AI$^2$ Robotics,
  \authorrefmark{19}Realman Robotics,
  \authorrefmark{20}Booster Robotics,
  \authorrefmark{21}DORA Community, \\
  \authorrefmark{22}Tsinghua University,
  \authorrefmark{23}Stanford University,
  \authorrefmark{24}University of California, Berkeley,
  \authorrefmark{25}PsiBot \\
}
\vspace{1mm}
\authorblockA{
  *\ Equal Contribution, 
  \dag\ Project Leaders,
  \ddag\ Corresponding Authors \\
  \texttt{\textcolor{blue}{shihan.wu.koorye@outlook.com, gcyao1@baai.ac.cn}} \\ 
}
\vspace{1mm}
\authorblockA{
  \textcolor{red}{\textit{\url{https://flagopen.github.io/RoboCOIN/}}}
}}

\twocolumn[{%
\renewcommand\twocolumn[1][]{#1}%
\maketitle
\thispagestyle{fancy}
\begin{center}
\centering
\includegraphics[width=\textwidth]{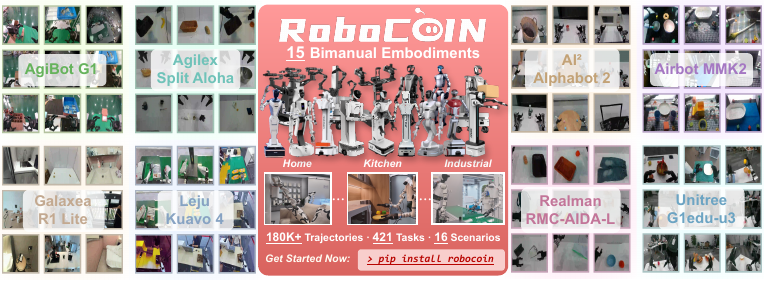}
\captionof{figure}{
Overview of our \textbf{RoboCOIN} dataset.
\textbf{RoboCOIN} is a large-scale, \textbf{multi-embodiment dataset explicitly built for bimanual robotic manipulation research}. It supports \textbf{\numembodiments} distinct robotic platforms, with \textbf{\numdemonstrationsabs} high-quality manipulation demonstrations spanning \textbf{\numtasks} structured bimanual tasks and \textbf{\numscenarios} diverse interactive scenarios.
The dataset is fully open to the research community, equipped with a ready-to-use one-click installation toolchain, and can be directly deployed via the simple command \textit{``pip install robocoin''}.
}
\label{fig:overview}
\end{center}
}]

\begin{abstract}
Despite the critical role of bimanual manipulation in endowing robots with human-like dexterity, large-scale and diverse datasets remain scarce due to the significant hardware heterogeneity across bimanual robotic platforms.
To bridge this gap, we introduce RoboCOIN, a large-scale multi-embodiment bimanual manipulation dataset comprising over \numdemonstrations demonstrations collected from \numembodiments distinct robotic platforms. 
Spanning \numscenarios diverse environments—including residential, commercial, and industrial settings—the dataset features \numtasks bimanual tasks systematically categorized by \numactions bimanual collaboration actions and \numobjects objects. 
A key innovation of our work is the hierarchical capability pyramid, which provides granular annotations ranging from trajectory-level concepts to segment-level subtasks and frame-level kinematics. 
Furthermore, we present CoRobot, an efficient data processing pipeline powered by the Robot Trajectory Markup Language (RTML), designed to facilitate quality assessment, automated annotation, and unified multi-embodiment and data management.
Extensive experiments demonstrate the effectiveness of RoboCOIN in enhancing the performance of various bimanual manipulation models across a wide spectrum of robotic embodiments.
The entire dataset and codebase are fully open-sourced, providing a valuable resource for advancing research in bimanual and multi-embodiment manipulation.
\end{abstract}

\IEEEpeerreviewmaketitle

\section{Introduction}

\textbf{Bimanual manipulation} is pivotal for enabling robots to perform complex, human-like tasks in unstructured real-world environments, ranging from manufacturing  \cite{pedersen2016robot} and home assistance  \cite{henschel2021makes} to logistics  \cite{echelmeyer2008robotics}. 
Critical to this ability is high-quality bimanual demonstration data, which underpins data-driven methods that require large-scale, diverse datasets, particularly those tailored for \textbf{multi-embodiment} scenarios, to generalize effectively across different robotic platforms \cite{xie2020deep,chen2022towards}.
Such data empowers models to master complex collaboration strategies, and develop robust control policies.

Despite significant progress in data-driven robotics, existing datasets remain constrained by limited diversity and structure, particularly for bimanual tasks and multi-embodiment learning.
The hardware heterogeneity across bimanual platforms impedes the collection of large-scale, diverse datasets  \cite{wang2025rethinking}. 
In addition, current benchmarks primarily supply low-level action trajectories while neglecting the structural learning of the manipulation process  \cite{o2024open}. 
Such limitations hinder the development of robotic platforms that can support learning systems capable of generalizing to novel environments \cite{liu2020skill, yang2024pushing}.

To bridge this gap, we propose \textbf{RoboCOIN}, a comprehensive \textbf{multi-embodiment bimanual manipulation} dataset comprising over \numdemonstrations demonstrations across \numtasks distinct bimanual tasks, integrated with \numscenarios diverse real-world scenarios.
As depicted in Figure \ref{fig:overview}, RoboCOIN is explicitly collected from \numembodiments diverse robotic platforms, which include three distinct bimanual configurations: bimanual robots, half-humanoid robots, and humanoid robots. 
These platforms are outfitted with a suite of cameras \textit{(e.g. head, wrist, etc.)} to provide diverse perspectives for comprehensive observation. 
They are also equipped with both parallel grippers and dexterous hands, which encompass a variety of configurations for heterogeneous bimanual embodiments.
All data are acquired via human teleoperation \cite{darvish2023teleoperation} to ensure high quality and diversity, accompanied by detailed language annotations.
Besides, a two-dimensional taxonomy based on action collaboration and object flexibility structures demonstrations to enable systematic task design and progressive skill acquisition.

To enable effective structural learning across diverse embodiments, RoboCOIN introduces a \textbf{Hierarchical Capability Pyramid} with three structured annotation levels:
\textit{i) Trajectory-level} annotations capture global concepts and task objectives for holistic planning; 
\textit{ii) Segment-level} annotations decompose tasks into executable subtasks with temporal alignment for structured reasoning; 
and \textit{iii) Frame-level} annotations provide dense details, including kinematic states and action labels for precise control. 
This multi-resolution pipeline supports learning from high-level conceptual understanding to low-level control, facilitating advanced reasoning and generalization in bimanual manipulation.

Additionally, we introduce \textbf{CoRobot}, a data processing pipeline for efficient development and deployment of RoboCOIN, which integrates three key components: 
\textit{i)} \textit{A Robot Trajectory Markup Language (RTML)} evaluator that validates trajectory properties, to ensure physical and semantic integrity across diverse robotic platforms; 
\textit{ii)} \textit{A Hierarchical Annotation Pipeline} to automatically generate multi-level annotations from raw demonstration data; 
and \textit{iii)} \textit{A Unified Robotic Interface} which extends LeRobot \cite{cadene2024lerobot} framework to provide unified multi-embodiment control, data management, and deployment capabilities for bimanual manipulation research.
This integrated infrastructure ensures consistent data quality across diverse hardware platforms while providing the necessary tools for scalable multi-embodiment learning.

Comprehensive evaluations across diverse models and robotic platforms validate notable performance improvements for bimanual manipulation tasks.
In the simulation environment, relative gains reach \textbf{75.7\%} on \textit{ARX-X5} and \textbf{212.9\%} on \textit{Franka Emika Panda} for bimanual manipulation tasks.
Real-world experiments further show substantial gains from the hierarchical pyramid, with \textbf{53.6\%} in-distribution and \textbf{155.6\%} out-of-distribution relative improvements.
Additionally, RTML filters out \textbf{35.3\%} of low-quality trajectories, yielding a \textbf{22.2\%} relative performance boost, demonstrating the importance of high-quality demonstration data.

The key contributions of this work are as follows:
\begin{itemize}
  \item 
  \textbf{A Large-scale Dataset.} 
  We introduce RoboCOIN, a large-scale  multi-embodiment bimanual dataset comprising over \numdemonstrations demonstrations across \numtasks tasks, collected from \numembodiments distinct robotic platforms, with hierarchical annotations from trajectory to frame levels.
  \item \textbf{An Efficient Data Processing Pipeline.}
  We develop a unified data processing pipeline named CoRobot, including RTML-based assessment, an automated annotation pipeline, and a platform for unified multi-embodiment dataset management and robot deployment.
  \item \textbf{Comprehensive Evaluations.}
  We perform extensive experiments across a wide spectrum of robotic platforms, demonstrating the effectiveness of RoboCOIN in enhancing bimanual manipulation models.
\end{itemize}

\section{Related Work}

\begin{figure*}[t]
\centering
\includegraphics[width=\textwidth]{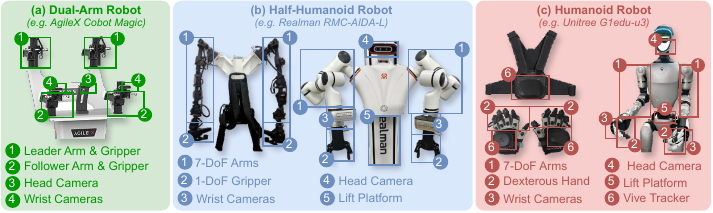}
\caption{
\textbf{Data collection platforms of our RoboCOIN}.
\textit{(a) Dual Robot (e.g., AgileX Cobot Magic). }
A bimanual robot with two 6-DoF arms and parallel grippers, capable of performing complex bimanual tasks.
\textit{(b) Half-Humanoid Robot (e.g., Realman RMC-AIDA-L). }
Left: wearable teleoperation device for human demonstration, right: bimanual robot with parallel grippers.
\textit{(c) Humanoid Robot (e.g., Unitree G1edu-u3). }
\textit{Left:} Wearable motion capture device for human demonstration. \textit{Right:} Humanoid robot with advanced dexterous manipulation capabilities. Head and wrist cameras provide multi-view visual observations.
}
\label{fig:robot}
\end{figure*}

\begin{table*}[h!]
\centering
\setlength{\tabcolsep}{6pt}
\caption{Robotic platforms used in RoboCOIN dataset collection.}
\label{tab:robots}
\begin{tabular}{lccccc}
\toprule
\textbf{Type} & \textbf{Name} & \textbf{Arm DoF} & \textbf{Camera Configuration} & \textbf{Gripper Type} & \textbf{Teleoperation Method} \\
\midrule
\multirow{3}{*}{\textbf{Dual-Arm}} 
& Agilex Cobot Magic \cite{cobot} & 2$\times$6 & Head + Wrist & Parallel Gripper & Isomorphic Arm \\
& Agilex Split ALOHA \cite{cobot} & 2$\times$6 & Head + Wrist & Parallel Gripper & Isomorphic Arm \\
& Galaxea R1 Lite \cite{galaxear1} & 2$\times$6 & Head + Wrist & Parallel Gripper & Isomorphic Arm \\
\midrule
\multirow{10}{*}{\textbf{Half-Humanoid}} 
& Realman RMC-AIDA-L \cite{realman} & 2$\times$7 & Head + Wrist & Parallel Gripper & Exoskeleton \\
& Agibot G1 \cite{agibot_g1} & 2$\times$7 & Head + Wrist + Back & Parallel Gripper & Virtual Reality \\
& AI$^2$ AlphaBot 2 \cite{ai2alpha} & 2$\times$7 & Head + Wrist + Chest & Dexterous Hand & Virtual Reality / Isomorphic Arm \\
& AI$^2$ AlphaBot 1s \cite{ai2alpha} & 2$\times$7 & Head + Wrist + Chest & Dexterous Hand & Virtual Reality / Isomorphic Arm \\
& Galbot G1 \cite{galbot} & 2$\times$7 & Head + Wrist & Parallel Gripper & Motion Capture \\
& Tianqing A2 \cite{tianqinga2} & 2$\times$7 & Head + Wrist + Back & Parallel Gripper & Virtual Reality \\
& Realman Rs-02 \cite{realman} & 2$\times$7 & Head + Wrist & Parallel Gripper & Exoskeleton \\
& Realman Rs-01 \cite{realman} & 2$\times$7 & Head + Wrist & Parallel Gripper & Exoskeleton \\
& Airbot MMK2 \cite{aitbotmmk2} & 2$\times$6 & Head + Wrist + Third-Person & Dexterous Hand & Virtual Reality \\
& Leju Kuavo 4 LB \cite{lejukuavo} & 2$\times$7 & Head + Wrist & Dexterous Hand & Virtual Reality \\
\midrule
\multirow{2}{*}{\textbf{Humanoid}} & Leju Kuavo 4 Pro \cite{lejukuavo} & 2$\times$7 & Head + Wrist & Dexterous Hand & Virtual Reality \\
& Unitree G1edu-u3 \cite{unitreeg1} & 2$\times$7 & Head + Wrist & Dexterous Hand & Motion Capture / Exoskeleton \\
\bottomrule
\end{tabular}
\end{table*}

\noindent\textbf{Robotic Learning Datasets.}
The evolution of robot learning has been significantly driven by the availability of diverse and scalable demonstration datasets. 
Early robot learning, limited by hardware, relied on simulation datasets \textit{(e.g., Meta-world    \cite{yu2020meta}, LIBERO    \cite{liu2023libero}, CALVIN    \cite{mees2022calvin})} but faced sim-to-real transfer challenges    \cite{zhao2020sim}.
Open-X-Embodiment    \cite{o2024open} aggregates real-world multi-embodiment data for better generalization but is limited to single-arm tasks, hindering its use for complex real-world bimanual interactions.
The $\pi_0$ dataset    \cite{black2024pi_0} provides extensive long-horizon bimanual trajectories but is proprietary and closed-source, restricting access for most researchers.
Recent datasets \textit{(e.g. AgiBot World    \cite{bu2025agibot}, Galaxea Open-World    \cite{jiang2025galaxea})} address the need for large-scale open-world bimanual data. 
AgiBot World offers large-scale, industrial-grade multi-scenario data, while Galaxea Open-World provides high-quality, fine-grained annotations from a unified mobile bimanual platform. 
However, these datasets often rely on a single robot embodiment, limiting their multi-embodiment applicability.
Generally, existing datasets lack multi-embodiment bimanual data with structured hierarchical annotations, constraining the development of generalizable bimanual manipulation models.

\noindent\textbf{Robotic Learning Policies.} 
Robot learning policies have advanced from specialized small-scale models to large-scale generalist systems, particularly for bimanual manipulation.
Early methods \textit{(e.g. ACT    \cite{zhao2023learning}, Diffusion Policy    \cite{chi2023diffusion})} performed adequately on specific tasks with small datasets but were limited by data scale and diversity, driving the development of generalist Vision-Language-Action (VLA) models.
For example, Octo    \cite{team2024octo} (trained on Open X-Embodiment dataset    \cite{o2024open}) supports language and goal-image conditioning, enabling zero-shot adaptation to new tasks or robots.
OpenVLA    \cite{kim2024openvla}, by contrast, uses an LLM-based autoregressive architecture, tokenizing continuous actions into discrete representations for LLM compatibility.
For bimanual manipulation specifically, RDT \cite{liu2024rdt} stands as a pioneering diffusion-based foundation model, which learns from diverse modalities via a scalable transformer and an interpretable unified action space tailored to bimanual collaboration.
$\pi_0$    \cite{black2024pi_0} is a model specifically designed for bimanual manipulation, using a flow-matching architecture, combining a VLM for perception and reasoning with an action expert for high-precision continuous motor commands.
GR00T-N1    \cite{gr00tn1_2025} is dedicated to bimanual humanoid manipulation and introduced a dual-system architecture for humanoids, with GR00T-N1.5    \cite{gr00tn15} adding upgrades like an improved VLM and FLARE objective.
Specialized models \textit{(e.g. Agibot GO-1    \cite{bu2025agibot}, Galaxea G0    \cite{jiang2025galaxea}, RoboBrain-X0    \cite{robobrainx0,ji2025robobrain})} enable effective single-embodiment bimanual control via multi-stage training and hierarchical systems.
All these models have their respective advantages, while the models designed for precise bimanual control rely on high-quality and diverse bimanual manipulation data. 
Despite their scale, these bimanual-related models lack structured task hierarchy understanding, limiting their reasoning ability for complex bimanual tasks.

\begin{figure*}[t]
\centering
\includegraphics[width=\textwidth]{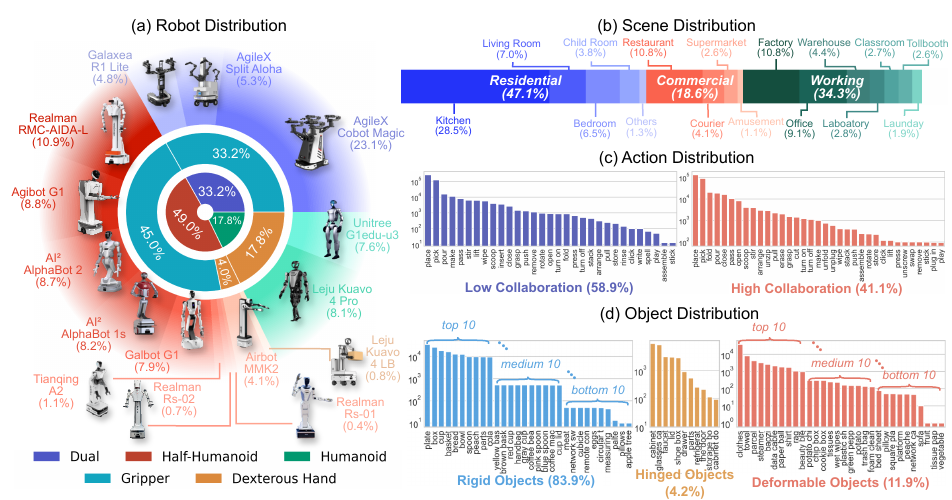}
\caption{\textbf{Overview statistics of the RoboCOIN dataset.}
RoboCOIN incorporates {(a)} \numembodiments distinct robotic platforms, including bimanual, half-humanoid, and humanoid robots with grippers and dexterous hands; 
{(b)} \numscenarios diverse environments such as residential, commercial, and working scenes; 
{(c)} \numactions action types categorized by collaboration levels; 
and {(d)} \numobjects object types spanning rigid, articulated, and deformable categories.}
\label{fig:overview_stats}
\end{figure*}

\section{RoboCOIN Dataset}

RoboCOIN is introduced as a pioneering \textbf{multi-embodiment bimanual manipulation} dataset, featuring a comprehensive suite of hierarchical annotations designed to facilitate advanced robot learning. 
It integrates \numembodiments diverse robotic platforms, \numdemonstrationsabs high-quality demonstrations, \numtasks structured bimanual tasks, and \numscenarios real-world scenarios. 
A Hierarchical Capability Pyramid is introduced to span three levels of annotations, facilitating multi-resolution learning from high-level concepts to low-level control.

\begin{figure*}[t]
\centering
\includegraphics[width=\textwidth]{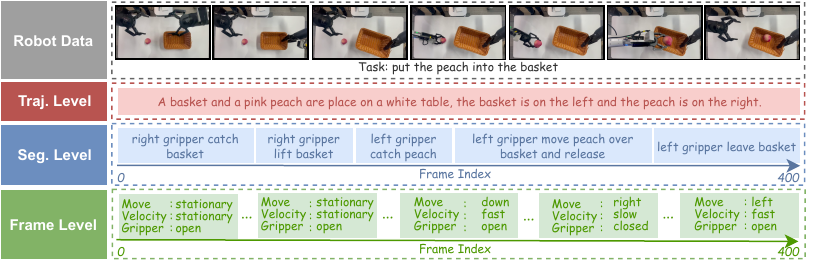}
\caption{\textbf{Hierarchical capability pyramid of RoboCOIN}. 
\textbf{Trajectory-level Annotations} define the global concepts; 
\textbf{Segment-level Annotations} decompose the task into executable subtasks; 
and \textbf{Frame-level Annotations} provide dense low-level details such as motion trajectories and gripper states. All annotations are temporally synchronized to form a cohesive data structure.
}
\label{fig:annotation} 
\end{figure*}

\subsection{Data Collection and Storage}

We use a diverse set of \numembodiments robotic platforms for comprehensive data acquisition, encompassing bimanual, half-humanoid, and humanoid configurations to ensure broad applicability across different bimanual manipulation scenarios.
Figure \ref{fig:robot} illustrates three representative platforms: bimanual robots \textit{(e.g., AgileX Cobot Magic    \cite{cobot})}, half-humanoid robots \textit{(e.g., Realman RMC-AIDA-L    \cite{realman})}, and humanoid robots \textit{(e.g., Unitree G1edu-u3    \cite{unitreeg1})}.
We employs teleoperation to ensure high-quality data collection, utilizing methods such as leader-follower isomorphic arms, exoskeletons, and motion capture systems, etc.
The platforms are equipped with a comprehensive suite of cameras, including head-mounted and wrist-mounted (chest, back, and third-person cameras for some platforms) to provide multi-view visual observations.
Featuring both parallel grippers and dexterous hands, kinematic data is recorded at high frequency to capture joint angles, end-effector poses, and gripper states, supporting accurate monitoring of robot motion, object interactions, and the diverse configuration characteristics of different platforms.
All kinematic measurements adhere to standardized conventions to ensure consistency across diverse robotic platforms, with joint states measured in radians and gripper states normalized between 0 and 1. 
The end-effector state includes position and orientation, where distances are expressed in meters, orientation is represented using Euler angles in radians, and all parameters follow a unified right-handed coordinate system with x-forward, y-left, and z-up.
Strict temporal synchronization across all data streams is maintained through timestamp alignment, ensuring consistency between visual observations and kinematic states.
Table \ref{tab:robots} provides a detailed overview of the robotic platforms used in RoboCOIN, including their arm degrees of freedom (DoF), camera configurations, gripper types, and teleoperation methods.
This diverse set of platforms ensures that the collected data covers a wide range of bimanual operational scenarios, from basic dual-arm control to complex humanoid bimanual interaction, laying a solid foundation for cross-embodiment bimanual learning.

\subsection{Statistics and Taxonomy}
\label{sec:stats_taxonomy}

\noindent\textbf{Robotic Platforms.}
RoboCOIN is explicitly designed as a multi-embodiment foundation to ensure broad applicability across diverse robotic platforms. Figure \ref{fig:overview_stats}(a) shows the distribution of robotic platforms in RoboCOIN, encompassing dual-arm, half-humanoid, and humanoid types. The \numembodiments distinct platforms offer rich morphological diversity for complex bimanual manipulation, with each type featuring unique characteristics tailored to different bimanual manipulation scenarios:
\begin{itemize}
\item \textbf{Dual-arm Robots (33.2\%).} 
Comprising 3 distinct models, these platforms focus on core dual-arm manipulation capabilities, making them suitable for fundamental bimanual tasks while maintaining low hardware costs.
\item \textbf{Half-humanoid Robots (49.0\%).} 
As the mainstream architecture in RoboCOIN, this category includes 10 distinct platforms. Among them, 45\% are equipped with standard parallel grippers, while 4\% feature dexterous hands, balancing human-like movement patterns with practical hardware requirements.
\item \textbf{Humanoid Robots (17.8\%).} 
Consisting of 2 platforms, these robots are equipped with fully dexterous hands, delivering advanced human-like manipulation capabilities.
\end{itemize}
This multi-embodiment platform portfolio ensures support for a wide spectrum of bimanual tasks, ranging from basic collaboration to complex dexterous manipulation.

\noindent\textbf{Scenarios.}
As shown in Figure \ref{fig:overview_stats}(b), data collection spans \numscenarios distinct scenarios, categorized into three main environments to ensure broad applicability:
\begin{itemize}
\item \textbf{Residential Scenarios (47.1\%).} As the primary scenario, it covers daily bimanual tasks such as cooking, cleaning, and home organization, which are closely aligned with human daily life scenarios.
\item \textbf{Commercial Scenarios (18.6\%).} Including restaurant, courier, supermarket, and amusement scenarios, it focuses on bimanual collaboration tasks \textit{(e.g., order picking, goods handling)} that require frequent interaction between both hands.
\item \textbf{Working Scenarios (34.3\%).} Covering factory, office, laboratory, and other scenarios, it supports complex bimanual operations such as equipment debugging and material handling.
\end{itemize}
Each category includes finer-grained scenarios \textit{(e.g., commercial settings cover restaurants, courier services, supermarkets, etc.)}, covering diverse real-world applications.

\noindent\textbf{Task Taxonomy.} 
RoboCOIN employs a two-dimensional task taxonomy that organizes manipulation scenarios along action collaboration (Figure \ref{fig:overview_stats}(c)) and object flexibility (Figure \ref{fig:overview_stats}(d)). 
Specifically, \numactions action patterns are categorized by the degree of bimanual collaboration required, including:
\begin{itemize}
\item \textbf{Low-collaboration Tasks (58.9\%).} 
Tasks where the two arms operate largely sequentially, with minimal synchronous interaction between the two arms.
\item \textbf{High-collaboration Tasks (41.1\%).} 
Tasks featuring partial or fully parallel arm movements, requiring close synchronous cooperation between the two arms.
\end{itemize}
To ensure the uniqueness of each action’s meaning and avoid terminology ambiguity, we strictly guarantee the uniqueness of each action’s definition, and a synonym mapping mechanism has been designed to standardize terminology usage.
The details can be founed in \textbf{Appendix \ref{appendix:action_list}}.

Similarly, \numobjects objects are classified along a spectrum of mobility, which can be divided into three types:
\begin{itemize}
\item \textbf{Rigid Objects (83.9\%).} 
Objects with fixed poses that do not undergo shape changes.
\item \textbf{Hinged Objects (4.2\%).} 
Objects with articulated parts that can move relative to each other.
\item \textbf{Deformable Objects (11.9\%).} 
Objects that can undergo significant shape changes across most parts.
\end{itemize}
This integrated framework facilitates the creation of diverse and incrementally challenging tasks that support skill development across different robotic platforms and real-world settings.

\subsection{Hierarchical Capability Pyramid}

\begin{table*}[t]
\setlength{\tabcolsep}{4.5pt}
\centering
\caption{\textbf{Qualitative comparison of existing real-world datasets for robot manipulation.}
    All data is drawn from the original paper or the RoboMIND paper.
    N/A indicates that the dataset does not provide explicit annotations for the corresponding attribute.
    \textcolor{gray}{$^{\dag}$ Not a dataset in itself, but an {aggregation} of existing datasets.}
}
\label{tab:dataset_comparison}
\begin{tabular}{lccccccccc}
\toprule
\textbf{Dataset} & \textbf{Arm} & \textbf{Embodiments} & \textbf{Trajectories} & \textbf{Tasks} & \textbf{Actions} & \textbf{Dexterous} & \textbf{Annotation} & \textbf{Collection Method} \\
\midrule
Pinto and Gupta   \cite{pinto2016supersizing} & Dual & 1 & 50k & N/A & 1 & \ding{55}  & No            & Scripted                   \\
RoboNet   \cite{dasari2019robonet}         & Single      & 1 & 162k & N/A  & N/A & \ding{55}  & No            & Scripted                   \\
MT-Opt   \cite{kalashnikov2021mt} & Single & 1 & 800k & 12 & 1 & \ding{55}  & No            & Scripted        \\
BridgeData   \cite{ebert2021bridge}        & Single      & 1 & 7.2k & 71   & 4   & \ding{55}  & No            & Human Teleoperation        \\
BC-Z   \cite{jang2022bc}                   & Single      & 1 & 26k  & 100  & 3   & \ding{55}  & No            & Human Teleoperation        \\
RH20T   \cite{fang2023rh20t}               & Single      & 1 & 13k  & 140  & 33  & \ding{55}  & No            & Human Teleoperation        \\
RoboSet   \cite{bharadhwaj2024roboagent}   & Single      & 1 & 98k  & 38   & 6   & \ding{55}  & No            & 30\% Human / 70\% Scripted \\
BridgeData V2   \cite{walke2023bridgedata} & Single      & 1 & 60k  & N/A  & 13  & \ding{55}  & No            & 85\% Human / 15\% Scripted \\
DROID   \cite{khazatsky2024droid}          & Single      & 1 & 76k  & N/A  & 86  & \ding{55}  & No            & Human Teleoperation        \\
\textcolor{gray}{Open X-Embodiment$^{\dag}$}   \cite{o2024open} & \textcolor{gray}{Single+Dual} & \textcolor{gray}{22} & \textcolor{gray}{1.4M} & \textcolor{gray}{160k} & \textcolor{gray}{217} & \textcolor{gray}{\ding{55}}  & \textcolor{gray}{No}            & \textcolor{gray}{Dataset Aggregation}        \\
RoboMIND   \cite{wu2024robomind}           & Single+Dual & 4 & 107k & 479  & 38  & \checkmark & Flat          & Human Teleoperation        \\
AgiBot World Beta   \cite{bu2025agibot}    & Dual        & 1 & 1M   & 217  & 87 & \ding{55}  & Flat          & Human Teleoperation        \\
Open Galaxea   \cite{jiang2025galaxea}     & Dual        & 1 & 50k  & 150  & 58 & \checkmark & Flat          & Human Teleoperation        \\
\rowcolor{red!10}
\textbf{RoboCOIN (Ours)}                                & \textbf{Dual}        & \ \textbf{\numembodiments} & \numdemonstrationsabs & \numtasks & \numactions & \textbf{\checkmark} & \textbf{Hierarchical}  & Human Teleoperation        \\
\bottomrule
\end{tabular}
\end{table*}

As illustrated in Figure \ref{fig:annotation}, the hierarchical capability pyramid in RoboCOIN encompasses three levels of structured annotations: \textit{trajectory-level, segment-level, and frame-level}, enabling multi-resolution learning from high-level conceptual understanding to low-level control.

\noindent\textbf{Trajectory-level Concepts.}
Trajectory-level annotations capture task concepts by characterizing the overall scene configuration. 
This involves a description of the scene, detailing the environment settings and the placement of objects. 
The resulting unified representation enables global spatial and physical reasoning throughout the entire task sequence    \cite{zhang2025dreamvla, shao2021concept2robot}.

\noindent\textbf{Segment-level Subtasks.}
Segment-level annotations decompose tasks into specific subtasks, which may temporally overlap to accommodate dual-arm operations. 
Each segment is aligned with specific video frames and includes step-by-step instructions. 
Annotations also explicitly label exception cases, such as grasping failures, to support robust error handling. 
This structured decomposition facilitates learning of temporal reasoning and task planning    \cite{belkhale2024rt,shi2025hi}.

\noindent\textbf{Frame-level Kinematics.}
Frame-level annotations describe kinematics for each frame. 
This includes the motion parameters \textit{(e.g., direction, velocity, acceleration)} for both arm end-effectors, and the status of grippers or dexterous hands \textit{(such as open/close state or transitioning movements)}. 
This high-density kinematic data enables real-time intrinsic feedback control and precise motion execution    \cite{xue2025reactive}.

\subsection{Comparison with Existing Datasets}

As shown in Table \ref{tab:dataset_comparison}, RoboCOIN distinguishes itself from existing robot datasets with its open, generalizable design for complex bimanual manipulation. 
Most existing datasets are limited to single-arm robots \textit{(e.g., BridgeData V2 \cite{walke2023bridgedata}, DROID \cite{khazatsky2024droid})}, mixed-embodiment datasets with insufficient bimanual coverage \textit{(e.g., Open X-Embodiment \cite{o2024open}, RoboMIND \cite{wu2024robomind})}, or single-platform dual-arm datasets \textit{(e.g., AgiBot World Beta \cite{bu2025agibot}, Open Galaxea \cite{jiang2025galaxea})}. 
In contrast, RoboCOIN includes \numembodiments diverse platforms with parallel grippers and dexterous hands, plus a hierarchical capability pyramid with multi-level annotations.
As a unique, fully open-source dataset, it provides comprehensive multi-embodiment bimanual manipulation data, serving as an invaluable resource for advancing multi-embodiment bimanual manipulation research.


\section{CoRobot Data Processing Pipeline}

\begin{figure*}[t]
\centering
\includegraphics[width=\textwidth]{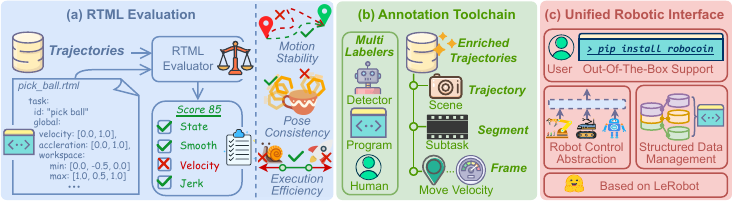}
\caption{
\textbf{Overview of the CoRobot data processing pipeline.}
(a) Robot Trajectory Markup Language (RTML) for automated trajectory validation.
(b) Annotation Pipeline for generating rich and hierarchical task descriptions.
(c) An out-of-the-box Unified Robotic Interface for unified robot control and multi-embodiment data management.
}
\label{fig:corobot}
\end{figure*}

For the efficient construction of the RoboCOIN dataset, we developed a CoRobot, an integrated data processing pipeline, as illustrated in Figure \ref{fig:corobot}. 
This pipeline encompasses three key components: \textit{i)} the Robot Trajectory Markup Language (RTML) for automated trajectory validation, \textit{ii)} a hierarchical annotation pipeline for generating rich and structured task descriptions, and \textit{iii)} a unified robotic interface for standardized robot control and multi-embodiment data management.

\subsection{Robot Trajectory Markup Language Evaluation}

High-quality data collection is challenged by distribution shifts arising from human operators' varying expertise, which deleteriously impact model performance.
To address this problem at the source, we propose the Robot Trajectory Markup Language (RTML), a domain-specific language designed to convert expert rules into standardized, machine-readable, and configurable constraints (Figure \ref{fig:corobot} (a)).
The design of RTML is based on three key principles for high-quality robot motion:
\begin{itemize}
  \item \textbf{Motion Stability.} Trajectories should be smooth and predictable, avoiding sudden changes that could lead to instability or inaccuracies.
  \item \textbf{Pose Consistency.} During the task execution phase, the robot's end-effector pose must meet task-specific constraints to ensure successful execution.
  \item \textbf{Execution Efficiency.} A trajectory must balance speed and precision, avoiding excessive haste or hesitation.
\end{itemize}
RTML is defined using the YAML format for readability and ease of use. 
It constrains trajectories from two perspectives: 
\textit{i)} global constraints that apply to the entire trajectory, defining motion characteristics including workspace boundaries, velocity limits, acceleration limits, and duration limits;
and \textit{ii)} local constraints that divide the trajectory into sequential phases \textit{(e.g., approach, grasp and place)}, defining override parameters and orientation tolerances for each phase.
Furthermore, an RTML evaluator is provided to automatically assess trajectory quality against the defined constraints.
The output includes detailed reports and an overall quality score, providing quantitative support for data selection and filtering.
The detailed specification of RTML can be found in \textbf{Appendix \ref{appendix:rtml}}.

\subsection{Annotation Pipeline}

We present a hierarchical annotation pipeline integrating detectors, rule-based programs, and human refinement, comprising three pipelines (Figure \ref{fig:corobot} (b)):
\begin{itemize}
\item \textbf{Trajectory-Level.} 
We use open-vocabulary detectors \cite{liu2024grounding,ren2024grounding} to localize objects and their spatial coordinates, which are fed into Large Language Models \cite{liu2024deepseek} to generate high-level scene descriptions, establishing semantic context for action understanding.
\item \textbf{Segment-Level.} 
Rule-based tools identify potential segmentation points \textit{(e.g., gripper opening/closing)} under human-specified guidance, with human annotators fine-tuning these points and filling in annotations.
\item \textbf{Frame-Level.} 
A cascaded sliding window framework extracts gripper kinematics. 
Taking displacement calculation as an example: the gripper's displacement at frame $t$ (denoted as $m_t$) is calculated by subtracting the 3D coordinates of the gripper at the $n$-th preceding frame ($x_{t-n}$) from its coordinates at the current frame ($x_t$), i.e., $m_t = x_t - x_{t-n} = (x_t^x - x_{t-n}^x, x_t^y - x_{t-n}^y, x_t^z - x_{t-n}^z)$.
For other kinematic indicators \textit{(e.g., velocity, acceleration)}, the sliding window calculation uses the previous window's output as input, enabling the extraction of higher-order motion features.
Finally, predefined thresholds convert these numerical kinematic metrics into consistent text labels, enabling fine-grained kinematic annotation.
\end{itemize}
After the automatic annotation process, human annotators review and refine the annotations to ensure accuracy.
Detailed implementations and analysis are provided in \textbf{Appendix C}.


\subsection{Unified Robotic Interface}

The heterogeneity of robot hardware interfaces, control protocols, and data representations presents a critical barrier to scalable multi-embodiment learning.
To alleviate these challenges, we introduce a Unified Robotic Interface that consolidates robot control and data management within a unified infrastructure (Figure \ref{fig:corobot}(c)).
Built on the LeRobot framework \cite{cadene2024lerobot}, the interface forms a robust, modular software stack with the following core functionalities:
\begin{itemize}
\item \textbf{Unified Control Abstraction.}
The interface integrates vendor-provided robot SDKs and is compatible with both ROS and ROS 2\cite{quigley2009ros,sciroboticsabm6074}, delivering a homogeneous control layer across diverse hardware.
It unifies device management and control paradigms, like camera configuration, coordinate alignment, and joint unit normalization, enabling consistent cross-embodiment operation.
\item \textbf{Structured Data Management.}
To enable efficient dataset reuse and flexible training data composition, the interface adopts an atomic data storage strategy that partitions datasets into minimal, reusable subsets tagged by embodiment, task type, and environment.
This structured approach allows researchers to easily query and assemble training data tailored to specific research needs.
\end{itemize}
By enabling one-click deployment through ``pip install robocoin'', the proposed interface effectively reduces engineering barriers for multi-embodiment learning research.
The system is fully open-source to support reproducibility and community-driven extensions.
\footnote{\url{https://github.com/FlagOpen/CoRobot}}


\begin{figure*}[t]
\centering
\includegraphics[width=\textwidth]{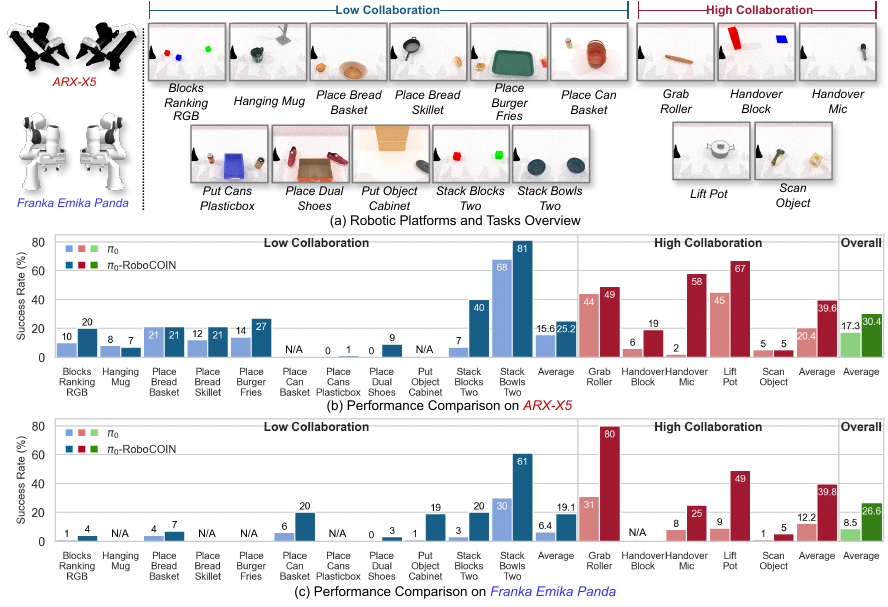}
\caption{
\textbf{Cross-embodiment transfer results on RoboTwin 2.0.}
(a) Our selected robotic platforms and bimanual tasks in the \textit{RoboTwin 2.0} environment.
(b) Performance comparison on \textit{ARX-X5}.
(c) Performance comparison on \textit{Franka Emika Panda}.
N/A indicates that the trajectory generation failed on this task.
}
\label{fig:cross_embodiment}
\end{figure*}

\section{Experiments and Analysis}

In this section, we evaluate the effectiveness of the RoboCOIN dataset through comprehensive experiments in both simulation and real-world environments.
We aim to answer the following research questions:
\begin{tcolorbox}[
    colback=white,
    colframe=black,
    boxrule=1pt,
    arc=0mm,
    left=2mm,
    right=2mm,
    top=1mm,
    bottom=1mm,
]
\textbf{\textit{Q1.}}
\textit{How does the RoboCOIN dataset enhance the multi-embodiment capability of VLA models?}

\textbf{\textit{Q2.}}
\textit{Does the hierarchical capability pyramid improve VLA models, and in what aspects?}

\textbf{\textit{Q3.}}
\textit{To what extent does the RTML contribute to improving data quality and model performance?}
\end{tcolorbox}

\subsection{Experiment Setup}

\noindent\textbf{Simulation Environment.}
The simulation experiment focuses on \textbf{\textit{Q1}}.
We conduct evaluations in the \textit{RoboTwin 2.0} environment \cite{mu2025robotwin,chen2025robotwin} using two distinct dual-arm platforms: \textit{ARX-X5} and \textit{Franka Emika Panda}, both of which are unseen embodiments in the training dataset.
We design 16 dual-arm manipulation tasks of varying difficulty, and as trajectory generation failed on a subset of tasks, we evaluate \textit{ARX-X5} on 14 tasks and \textit{Franka Emika Panda} on 11 tasks, respectively.
Visual observations are captured by \textit{Realsense D435} cameras mounted on the head and wrists, with a resolution of 320$\times$240.

\noindent\textbf{Real-World Environment.}
The real experiments focus on \textbf{\textit{Q2}} and \textbf{\textit{Q3}}, using two robotic platforms: the \textit{Realman RMC-AIDA-L} \cite{realman} and the \textit{Unitree G1edu-u3} \cite{unitreeg1}.
The \textit{Realman RMC-AIDA-L} is equipped with three \textit{Realsense D435 Cameras} and \textit{RMG24 Parallel Grippers} for visual perception and manipulation. 
On this platform, tasks are designed with varying complexity according to the difficulty of dual-arm collaboration and object flexibility, also covering out-of-distribution (OOD) test scenarios.
The \textit{Unitree G1edu-u3} is integrated with a \textit{Realsense D435 Camera} on the head and \textit{Realsense D405 Cameras} on the wrists, along with the \textit{Unitree Dex3-1 Three-fingered Dexterous Hand}. 
For this platform, two representative tasks are adopted to investigate how trajectory data quality affects policy learning.

\noindent\textbf{Representative Tasks.}
To better illustrate task execution pipelines and inherent challenges, we select several representative tasks from our experiments for detailed description, including both simulated and real-world settings:
\begin{itemize}
    \item \textbf{Simulation Lift Pot.}
    The robot grasps both handles of a pot with two arms and lifts it off the ground. The task succeeds when the pot is fully separated from the ground. The main challenge is highly synchronous bimanual collaboration to prevent tilting or dropping.
    \item \textbf{Simulation Place Dual Shoes.}
    The robot grasps two shoes at the same time and places them into a box one after another. The task succeeds when both shoes are completely inside the box. This task demands bimanual collaboration and accurate placement control.
    \item \textbf{Real-World Peach Drawer.}
    The robot uses one arm to place a peach into a drawer and the other to close the drawer. The task succeeds when the peach is inside the drawer and the drawer is fully closed. It requires precise asynchronous action execution.
    \item \textbf{Real-World Pass Bowl.}
    The robot hands a bowl from one arm to another in mid-air and places it at the target position. The task succeeds when the bowl is stably placed at the target. The key difficulty lies in stable handover and tight bimanual collaboration.
\end{itemize}
The detailed simulation and real-world experimental settings and visual sequences can be found in \textbf{Appendix \ref{appendix:tasks}}.

\noindent\textbf{Evaluated VLA Policies.}
To comprehensively validate the generalization capability of our proposed framework across diverse robotic platforms, we evaluate two state-of-the-art VLAs specifically designed for bimanual manipulation tasks:

\begin{itemize}
  \item \textbf{$\pi_0$}   \cite{black2024pi_0}. 
  A flow-matching-based VLA model with a large vision-language model and a action expert network.
  \item \textbf{GR00T N1.5}   \cite{gr00tn15}.
  A dual-system VLA model for humanoid robots, featuring a vision-language model and a future state alignment objective.
\end{itemize}

To enable parameter-efficient adaptation, we fine-tuned both models on data from all experimental tasks (with 50 trajectories per task) in a joint training manner: $\pi_0$ using LoRA \cite{hu2022lora}, while \textit{GR00T N1.5} was adapted via partial fine-tuning on only its diffusion and projector components.
For $\pi_0$, we used the AdamW optimizer with a learning rate of \(2.5\times10^{-5}\), batch size 32, 30000 training steps.
For \textit{GR00T N1.5}, we also employed AdamW optimizer, with a learning rate of \(1\times10^{-4}\), batch size 24, 10000 training steps.

\begin{figure}[t]
\centering
\includegraphics[width=0.49\textwidth]{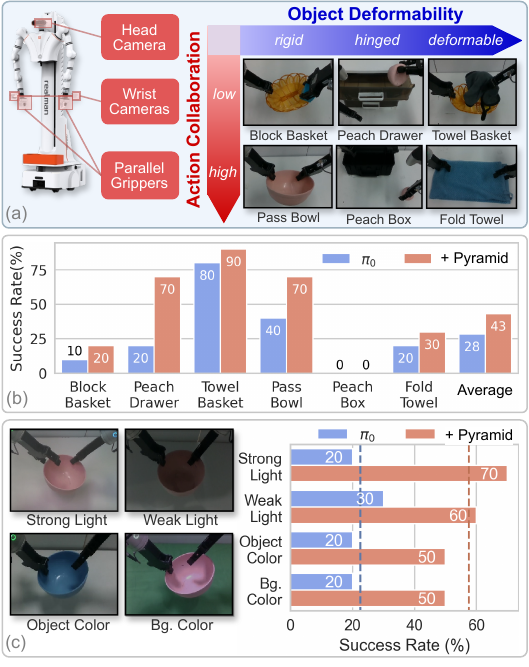}
\caption{\textbf{Results of Realman RMC-AIDA-L over the $\pi_0$ baseline}. 
It follows an multi-dimensional task design (a), with the in-distribution (b) and out-of-distribution results (c) showing the success rates with and without HAI.
}
\label{fig:hai_results}
\vspace{-8pt}
\end{figure}

\subsection{Impact of Multi-Embodiment Adaptability}

To answer \textbf{\textit{Q1}}, we design a cross-embodiment transfer experiment based on the simulation \textit{RoboTwin 2.0} environment with the dual-arm \textit{ARX-X5} and \textit{Franka Emika Panda} robots as the target embodiments.
It compares the fine-tuning performance of the original model and the model post-trained on the RoboCOIN dataset.
\begin{itemize}
  \item $\pi_0$: 
  The baseline model fine-tuned from the original $\pi_0$ using selected \textit{RoboTwin 2.0} tasks.
  \item $\pi_0$\textbf{-RoboCOIN}: 
  The optimized model post-trained on the random sample RoboCOIN-10K subset, then fine-tuned on selected \textit{RoboTwin 2.0} tasks.
\end{itemize}

As illustrated in Figure \ref{fig:cross_embodiment}, $\pi_0$\textbf{-RoboCOIN} achieves consistent and substantial performance gains over $\pi_0$ across nearly all tasks on both robotic platforms.
On the \textit{ARX-X5} platform, $\pi_0$\textbf{-RoboCOIN} improves the average success rate from \textbf{17.3\%} to \textbf{30.4\%}, representing a relative improvement of \textbf{75.7\%}.
On the \textit{Franka Emika Panda} platform, the average success rate rises from \textbf{8.5\%} to \textbf{26.6\%}, delivering a \textbf{212.9\%} relative improvement.
Notably, high-collaboration tasks benefit significantly more from RoboCOIN than low-collaboration tasks.
On \textit{ARX-X5}, low-collaboration tasks improve moderately from \textbf{15.6\%} to \textbf{25.2\%} \textit{(a relative gain of 61.5\%)}, whereas high-collaboration tasks surge from \textbf{20.4\%} to \textbf{39.6\%}, with a remarkable relative improvement of \textbf{94.1\%}.
On \textit{Franka Emika Panda}, the performance boost is even more prominent across both collaboration levels.
Notably, for challenging tasks such as \textit{``Place Dual Shoes''}, the success rate increases from \textbf{0\%} to \textbf{9\%} on \textit{ARX-X5} and from \textbf{0\%} to \textbf{3\%} on \textit{Franka Emika Panda}, turning previously unachievable tasks into feasible ones.
This observation underscores that multi-embodiment data in RoboCOIN endows the policy with strong, transferable high-collaboration manipulation priors and effectively enhances the generalization of robotic policies to unseen hardware platforms, especially for complex high-collaboration bimanual manipulation tasks.

\begin{figure*}
\centering
\includegraphics[width=\textwidth]{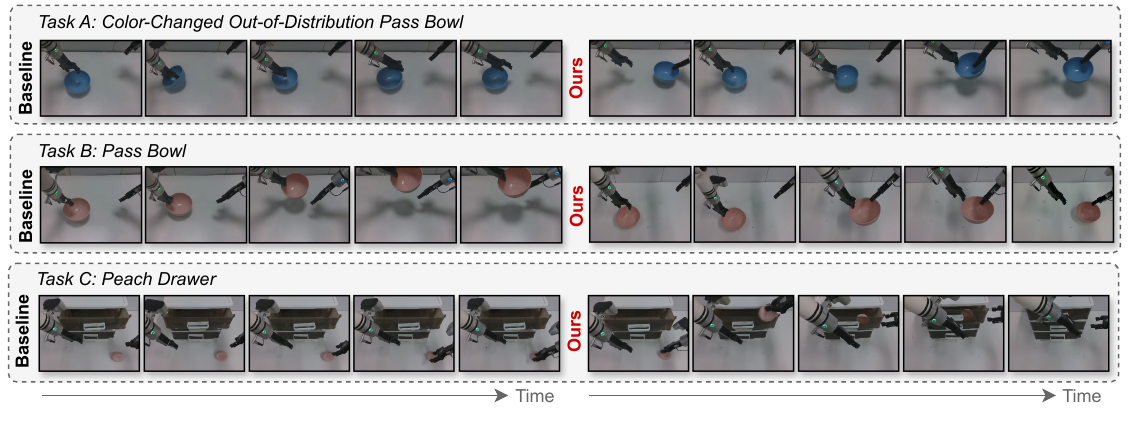}
\caption{\textbf{Case study} of the impact of RTML on data quality and model performance.}
\label{fig:case_analysis}
\end{figure*}

\subsection{Impact of Hierarchical Capability Pyramid}

To address \textbf{\textit{Q2}}, we investigate how the hierarchical capability pyramid contributes to robotic manipulation, with experiments conducted on the \textit{Realman RMC-AIDA-L} platform using the $\pi_0$.
We design two structurally consistent pipelines for controlled comparison:
\begin{itemize}
    \item \textbf{$\pi_0$ Baseline.}
    The model takes raw visual observations, proprioceptive states, and standard natural language instructions as input, and directly outputs manipulation actions in an end-to-end manner.
    \item \textbf{$\pi_0$ + Pyramid.}
    A minimal, non-intrusive integration strategy that introduces the hierarchical capability pyramid without modifying the model architecture or parameters.
    Specifically, we enrich the language input with structured hierarchical task, subtask, and action descriptions, which are formed by fusing human instructions and automatically extracted motion cues from trajectory history.
\end{itemize}
This design ensures the augmented model can naturally incorporate structured task priors and motion knowledge from the hierarchical capability pyramid into its decision-making process, while maintaining strict consistency with the baseline pipeline for fair comparison. 
Detailed implementation details of the integration strategy are provided in \textbf{Appendix \ref{appendix:hai}}.

\noindent\textbf{In-Distribution Performance.}
Figure \ref{fig:hai_results} (a) shows our multi-level task suite, with complexity determined by dual-arm collaboration difficulty and object flexibility.
As observed in Figure \ref{fig:hai_results} (b), introducing the hierarchical capability pyramid leads to consistent and significant performance gains across all tasks, with an average success rate improvement from \textbf{28\%} to \textbf{43\%}, representing a \textbf{53.6\%} relative improvement.
Notably, the success rate of the complex task \textit{``Peach Drawer''} rises sharply from \textbf{20\%} to \textbf{70\%}, highlighting that the hierarchical structure is beneficial for long-horizon, highly coordinated tasks.

\noindent\textbf{Out-of-Distribution Robustness.}
We further evaluate model robustness by directly deploying in-distribution models to four OOD variants of the \textit{``Pass Bowl''} task, involving changes in lighting, background, and object appearance.
As shown in Figure \ref{fig:hai_results} (c), the benefit of the hierarchical capability pyramid remains strong under distribution shift.The average success rate across OOD tasks improves from \textbf{22.5\%} to \textbf{57.5\%}, representing a \textbf{155.6\%} relative improvement.
Compared to in-distribution results \textit{(40\% and 70\%)}, the baseline $\pi_0$ suffers a \textbf{43.8\%} relative performance drop, whereas the pyramid-augmented model only drops by \textbf{17.9\%}.
These results confirm that the hierarchical capability pyramid effectively improves robustness to environmental variations.

\noindent\textbf{Qualitative Case Study.}
To further illustrate the functional improvements brought by the hierarchical capability pyramid at different levels, we present three representative qualitative cases covering global scene understanding, subtask transition, and fine-grained manipulation, as shown in Figure \ref{fig:case_analysis}:
\begin{itemize}
\item \textbf{Color-changed Out-of-Distribution Pass Bowl (20\% $\to$50\%).}
The baseline fails to grasp the bowl under appearance distribution shift, while our approach completes grasping and task execution, demonstrating improved \textit{scene-level perception and generalization}.
\item \textbf{Pass Bowl (40\%$\to$70\%).}
The baseline gets stuck during arm handover, while our method achieves smooth subtask switching, showing enhanced \textit{segment-level transition capability}.
\item \textbf{Peach Drawer (20\%$\to$70\%).}
The baseline fails in fine-grained peach grasping, while our method performs stable manipulation, indicating improved \textit{frame-level fine manipulation ability}.
\end{itemize}
Together, these cases demonstrate that the hierarchical capability pyramid systematically enhances performance across scene-level perception, segment-level subtask transition, and frame-level fine-grained manipulation—directly aligning with its multi-level structural design and validating that the pyramid's hierarchical priors effectively address the key limitations of the baseline model at each corresponding level.

\begin{figure}[t]
\centering
\includegraphics[width=0.49\textwidth]{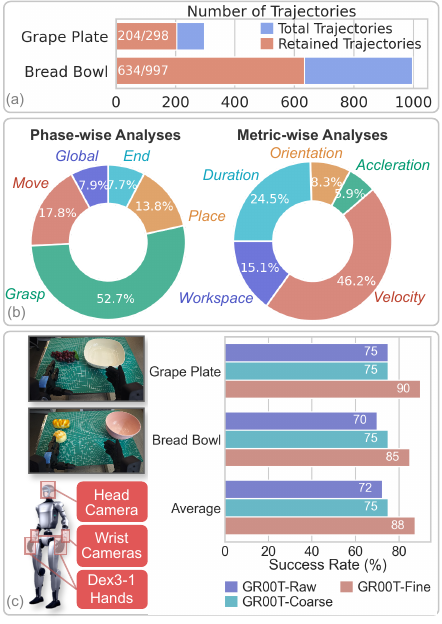}
\caption{
\textbf{Impact of RTML on data quality and model performance.}
(a) RTML filtering effect on two tasks.
(b) Failure cause analysis from phase-wise and metric-wise perspectives.
(c) Model performance improvement of \textit{Unitree G1edu-u3 + GR00T N1.5} under 3 fine-tuning settings.
}
\label{fig:rtml_results}
\end{figure}

\subsection{Impact of RTML}

To answer \textbf{\textit{Q3}}, we conduct experiments on two tasks using the \textit{Unitree G1edu-u3} platform.
As shown in Figure \ref{fig:rtml_results}(a), RTML filters out an average of \textbf{35.3\%} of low-quality trajectories across the two tasks, indicating that the original human demonstration data contains a large amount of inconsistent information that may interfere with effective model learning, verifying the necessity of data quality optimization.
Combined with phase-wise and metric-wise analyses (Figure \ref{fig:rtml_results}(b)), the key findings are summarized as follows:
\begin{itemize}
\item \textbf{Phase-level analysis.}
The grasping phase accounts for the highest proportion of failures (\textbf{52.7\%}), followed by the moving phase (\textbf{17.8\%}).
\item \textbf{Metric-level analysis.}
Velocity violation is the dominant cause of trajectory invalidity (\textbf{46.2\%}), with duration violation as the secondary factor (\textbf{24.5\%}).
\end{itemize}
Together, these observations identify critical failure modes and provide clear, actionable guidance for future demonstration collection and policy optimization.

To further investigate the impact of RTML on model performance, we further evaluated how RTML-filtered data affects policy learning by comparing three fine-tuning settings:
\begin{itemize}
\item \textbf{GR00T-Raw.} Fine-tuned on the original dataset without RTML filtering.
\item \textbf{GR00T-Coarse.} Fine-tuned on data filtered only by global RTML constraints.
\item \textbf{GR00T-Fine.} Fine-tuned on data filtered by both global and phase-wise constraints.
\end{itemize}
From the results in Figure \ref{fig:rtml_results}(c), the three configurations exhibit a clear progressive improvement in average success rates.
Compared to \textit{GR00T-Raw} with a success rate of \textbf{72\%}, \textit{GR00T-Coarse} achieves a modest \textbf{3\%} gain \textit{(4.2\% relative)}, while \textit{GR00T-Fine} achieves a significant \textbf{16\%} improvement \textit{(22.2\% relative)}.
This observation highlights the important impact of phase-level constraints on trajectory quality and policy learning.
These results demonstrate that RTML effectively enhances data quality by filtering out low-quality trajectories, and that this improvement in data quality translates into significant performance gains for the trained policies, especially when fine-grained, phase-specific constraints are applied.

\section{Conclusion, Limitations and Future Work}
In this work, we present \textbf{RoboCOIN}, a \textbf{large-scale multi-embodiment bimanual} dataset that unifies \textbf{\numembodiments robotic platforms}, over \textbf{\numdemonstrations demonstrations}, \textbf{\numtasks tasks}, and \textbf{\numscenarios scenarios} into a single benchmark. 
RoboCOIN bridges critical gaps in bimanual manipulation research by offering unprecedented hardware diversity and a hierarchical capability pyramid for structured learning. 
Complementing the dataset, we introduce \textbf{CoRobot}, an integrated pipeline powered by RTML for trajectory quality assessment, a semi-automatic annotation pipeline, and a unified platform for control and data management. 
Extensive experiments show that RoboCOIN significantly enhances multi-embodiment adaptability, hierarchical reasoning, and data quality for robotic learning, providing a solid foundation for future research in multi-embodiment bimanual manipulation.

While RoboCOIN serves as a significant catalyst for bimanual manipulation research, it is subject to several limitations that warrant further investigation. 
\textit{First,} reliance on teleoperation may introduce inherent operator biases, leading to inter-operator variability in trajectory patterns. 
\textit{Second,} manual annotation presents a scalability bottleneck due to its high labor costs and potential for subjectivity, which may impact data consistency. 
\textit{Third,} the current RTML framework depends heavily on expert-defined heuristics, potentially constraining its flexibility when applied to niche tasks or novel robotic platforms. 
These limitations may affect the overall robustness and generalization of the dataset and trained models, highlighting our commitment to ongoing research efforts aimed at enhancing data quality, annotation efficiency, and RTML adaptability in future iterations of RoboCOIN.

\bibliographystyle{plainnat}
\bibliography{references}

\newpage

\appendices

\section{Action and Object Taxonomy}
\label{appendix:action_list}

To ensure standardized data collection, consistent annotation, and robust policy learning across diverse robotic platforms, we propose a unified and hierarchical taxonomy for robotic manipulation actions.
The taxonomy is organized into 4 core categories based on \emph{functional purpose, physical interaction type, and environmental effect}, each representing a distinct class of robot behaviors.
The categories are defined as follows:
\begin{itemize}
    \item \textbf{General Manipulation.} Basic low-level primitives that describe universal, frequently used robot motions, including grasping, placement, pushing, pulling, and alignment.
    \item \textbf{Object State Change.} Actions that alter the physical state, shape, or configuration of an object, such as pressing, folding, twisting, shaking, and tearing.
    \item \textbf{Object Relation Change.} Operations that modify the spatial or structural relationship between multiple objects, including stacking, insertion, assembly, attachment, and exchange.
    \item \textbf{Task-Specific Actions.} High-level specialized operations for domain-specific scenarios, including cleaning, cooking, pouring, cutting, device control, and writing.
\end{itemize}
To enhance language grounding and improve model robustness to linguistic variations, each action verb is accompanied by semantically consistent synonyms.
These synonyms are used for textual data augmentation, enabling the model to better understand diverse natural language instructions and reduce annotation ambiguity.
The complete glossary of standardized action verbs, synonyms, and formal definitions is presented in Table~\ref{tab:action_glossary}, which serves as a foundational reference for data collection and annotation in the RoboCOIN dataset.

\begin{table*}[t]
\centering
\caption{Robotic Manipulation Action Glossary}
\label{tab:action_glossary}
\setlength{\tabcolsep}{20pt} 
\begin{tabular}{lccc}
\toprule
\textbf{Category} & \textbf{Verb} & \textbf{Synonyms} & \textbf{Definition} \\
\midrule
\multirow{9}{*}{General Manipulation}
& pick & grab, take, get & Grasp and lift an object from its position \\
& grasp & grip, clutch, hold & Firmly hold an object without positional change \\
& place & put, set, position & Put an object onto a surface or target location \\
& pass & hand over, give & Transfer an object between two hands/end-effectors \\
& rotate & twist, turn, revolve & Change the angular orientation of an object \\
& push & slide, shove & Move an object by applying forward force \\
& pull & drag, tug & Draw an object toward the agent \\
& lift & raise, hoist & Move an object upward to a higher position \\
& store & stow, deposit & Place an object inside a closed/semi-enclosed container \\
\midrule
\multirow{8}{*}{Object State Change}
& press & depress, push down & Apply continuous force to deform/activate an object \\
& click & tap, press briefly & Apply momentary force to actuate a button/switch \\
& open & unfold, unzip & Switch an object from closed to accessible state \\
& unzip & -- & Open a zippered fastener \\
& unfold & spread out, expand & Flatten a folded or compact object \\
& close & shut, zip & Switch an object from open to closed state \\
& fold & tuck, collapse & Bend an object into a compact/layered form \\
& unscrew & loosen, twist off & Release a lid/cap by counterclockwise rotation \\
\midrule
\multirow{9}{*}{Object Relation Change}
& stack & pile, heap & Arrange objects vertically in layers \\
& arrange & line up, organize & Place objects in an orderly spatial pattern \\
& insert & put in, slot & Place one object into another structure \\
& plug in & connect, attach & Insert an electrical plug into a power socket \\
& remove & pull out, extract & Take an object out of a space/connection \\
& unplug & disconnect & Remove an electrical plug from a socket \\
& assemble & construct, build & Combine parts into a functional unit \\
& stick & paste, attach & Affix an object to a surface via adhesion \\
& swap & exchange, switch & Exchange positions of two objects \\
\midrule
\multirow{13}{*}{Task-Specific Actions}
& wipe & scrub, clean & Clean a surface via rubbing motion \\
& erase & remove marks & Eliminate traces or residues from a surface \\
& rinse & wash, flush & Clean an object using flowing water \\
& turn on & activate, start & Power on an electrical/mechanical device \\
& turn off & deactivate, stop & Power off an electrical/mechanical device \\
& pour & transfer, spill & Move liquid from one container to another \\
& scoop & ladle, dig out & Lift material using a spoon or scooping tool \\
& stir & mix, blend & Homogenize liquid/particulate matter \\
& cut & chop, slice, split & Separate an object using a sharp tool \\
& play & perform & Operate a musical instrument \\
& make & build, create & Construct an object or prepare food \\
& spell & recite letters & Form words by sequencing individual characters \\
& write & inscribe, jot & Produce characters on a surface \\
\bottomrule
\end{tabular}
\end{table*}

\section{RTML Specification}
\label{appendix:rtml}

To formalize the constraints of robotic manipulation tasks in a structured and machine-interpretable manner, we design a \textbf{Robot Trajectory Markup Language (RTML)}, exemplified in Listing \ref{lst:rtml_example} for the "\textit{pull bowl storage bread}" task. 
RTML adopts a hierarchical constraint architecture that decouples \textbf{Global Constraints} and \textbf{Local Constraints}, enabling modular specification of task-wide kinematic limits alongside stage-specific spatial-temporal and kinematic requirements.

For \textbf{Global Constraints}, RTML defines universal kinematic limits that apply to the entire task execution, including upper bounds for linear velocity (maximum: 0.5 m/s, mean maximum: 0.3 m/s) and linear acceleration (maximum: 12.0 m/s²). 
These global constraints ensure basic safety and motion smoothness across all manipulation stages, preventing excessive kinematic values that may lead to mechanical instability or task failure.  

For \textbf{Local Constraints}, RTML partitions the overall task into sequential, semantically meaningful subtasks \textit{(e.g., \texttt{move\_bowl\_right}, \texttt{grasp\_long\_bread\_left}, \texttt{place\_long\_bread\_in\_bowl})}, each associated with a human-readable \texttt{match\_subtask} description to align with natural language task definitions. Each stage specifies multi-dimensional constraints tailored to the subtask's requirements: 
\begin{itemize}
\item \textbf{Workspace constraints} define the 3D Cartesian bounds for the active arm \textit{(e.g., the right arm's workspace for moving the pink bowl is restricted to \textit{[0.10, -0.40, 0.10]} to \textit{[0.25, -0.20, 0.30]} in meters)}, ensuring the robot operates within task-relevant spatial regions.
\item \textbf{Kinematic constraints} refine velocity limits \textit{(e.g., linear velocity mean maximum of 0.10 m/s for bowl movement)} with additional statistical constraints \textit{(e.g., standard deviation maximum of 0.08 m/s)} to control motion variability.
\item \textbf{Idle arm constraints} impose strict velocity limits on the non-active arm \textit{(e.g., left arm's linear velocity mean maximum of 0.05 m/s during right-arm bowl movement)} to avoid unintended motion interference.
\item \textbf{Orientation constraints} \textit{(e.g., angular mean deviation maximum of 0.8 rad and angular variance maximum of 0.15 for grasping long bread with the left arm)} regulate the pose stability of the end-effector for precise grasping.
\item \textbf{Temporal constraints} define the minimum and maximum duration for each stage \textit{(e.g., 2.0–8.0 seconds for bread grasping)} to ensure task execution efficiency and completeness.  
\end{itemize}
By structuring RTML in this hierarchical manner, we enable clear delineation between overarching task constraints and stage-specific requirements, facilitating data quality control during trajectory collection and providing a robust framework for downstream robotic learning algorithms to interpret and adhere to task constraints.

\section{Annotation Pipeline}
\label{appendix:annotation}

\begin{figure*}[h]
  \centering
  \includegraphics[width=\textwidth]{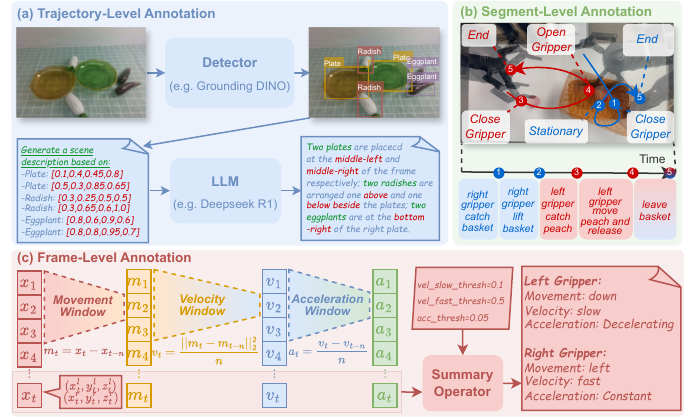}
  \caption{
    \textbf{Annotation Pipeline Design.}
    (a) Trajectory-Level Annotation Pipeline.
    (b) Segment-Level Annotation Pipeline.
    (c) Frame-Level Annotation Pipeline.
  }
  \label{fig:annotation_pipeline}
\end{figure*}

To systematically annotate the multi-faceted aspects of robotic manipulation trajectories, we develop a comprehensive annotation pipeline consisting of three hierarchical pipelines: \textit{Trajectory-Level Annotation, Segment-Level Annotation, and Frame-Level Annotation}, as illustrated in Figure \ref{fig:annotation_pipeline}.

\noindent\textbf{Trajectory-Level Annotation Pipeline.}
As shown in Figure \ref{fig:annotation_pipeline}(a), trajectory-level annotation provides the semantic basis for understanding robotic manipulation scenes by converting raw visual observations into human-readable natural language descriptions, including the following two key steps:
\begin{enumerate}
\item \textbf{Object Detection.}
A state-of-the-art detector (\textit{e.g., Grounding DINO}) processes visual frames to detect and localize key objects, outputting structured class labels and normalized bounding box coordinates.
The detector outputs structured information comprising object class labels and bounding box coordinates, formatted as numerical tuples that encapsulate the spatial distribution of each object (\textit{e.g., [0.1,0.4,0.45,0.8] for a plate, [0.3,0.25,0.5,0.5] for a carrot}).
\item \textbf{Scene Description Generation.}
This structured information is then passed to a large language model (\textit{e.g., Deepseek R1}), which translates low-level spatial data into coherent scene descriptions that summarize object positions, quantities, and spatial relationships, such as "\textit{Two plates are placed at the middle-left and middle-right of the frame respectively; two radishes are arranged one above and one below beside the plates; two eggplants are at the bottom-right of the right plate.}" 
\end{enumerate}
By lifting raw visual detections to high-level semantics, this step builds a structured contextual representation for understanding robot actions and supports further analysis of manipulation tasks.

\noindent\textbf{Segment-Level Annotation Pipeline.}
Figure \ref{fig:annotation_pipeline}(b) illustrates the segment-level annotation pipeline, which decomposes continuous robot manipulation trajectories into distinct action segments with fine-grained atomic sub-task labels.
The core approach is to first detect key temporal points that partition the trajectory, then assign semantic labels to each resulting segment.
Key temporal points for segment division are identified mainly based on robotic gripper states and task execution progress, including the following core types of temporal points:
\begin{itemize}
\item \textbf{Stationary state of the gripper.} 
This refers to the period when the gripper remains motionless during task execution, which usually indicates a transition between two consecutive atomic actions.
\item \textbf{Gripper opening and closing transitions.} 
The moments corresponding to the ``Open Gripper'' and ``Close Gripper'' states serve as critical division points.
For example, the moment the gripper starts to close marks the initiation of a grasping action, and the moment it starts to open indicates the initiation of a releasing action.
\item \textbf{Task end moment.} 
The point when the robot completes all operations of a specific sub-task or the entire task.
\end{itemize}
Notably, there are other potential temporal points that could be considered for segment division, such as sudden changes in gripper velocity or acceleration, or significant shifts in object positions.
After segmentation, each segment is labeled with a concise atomic sub-task description.
By decomposing continuous behaviors into semantically meaningful segments, this step improves the interpretability of robot trajectories and supports fine-grained task analysis.

\noindent\textbf{Frame-Level Annotation Pipeline.}
Figure \ref{fig:annotation_pipeline}(c) presents the frame-level annotation pipeline, which extracts fine-grained gripper kinematics at each frame along manipulation trajectories.
We use a cascaded window-based scheme to compute key motion metrics, as detailed below:
\begin{enumerate}
    \item \textbf{Movement Window.}
    Computes the gripper displacement vector $m_t$ as the coordinate difference between frame $t$ and frame $t-n$:
    $$
    m_t = x_t - x_{t-n},
    $$
    \item \textbf{Velocity Window.}
    Estimates frame-wise velocity $v_t$ from the L2 norm of consecutive displacement differences:
    $$
    v_t = \frac{\|m_t - m_{t-1}\|_2}{n}.
    $$
    \item \textbf{Acceleration Window.}
    Derives acceleration $a_t$ as the normalized change in velocity:
    $$
    a_t = \frac{v_t - v_{t-n}}{n}.
    $$
\end{enumerate}
It should be emphasized that the cascaded window-based computation pipeline described below is a typical example for kinematic information extraction, and the specific implementation can be flexibly adjusted according to actual needs.
This pipeline efficiently quantifies fine-grained gripper dynamics and supports robust trajectory quality assessment.

\noindent\textbf{Summary Operator.}
After computing frame-level kinematic metrics including displacement, velocity, acceleration, and other motion features, a summary operator maps continuous numerical values to discrete gripper state labels via predefined thresholds.
For example, given velocity threshold ``$\text{vel\_slow\_thresh}=0.1$'' and ``$\text{vel\_fast\_thresh}=0.5$'', the speed level is defined as follows:
$$
\text{Vel} = \begin{cases}
\text{Stationary}, & v < \text{vel\_slow\_thresh}, \\
\text{Slow}, & \text{vel\_slow\_thresh} \leq v \leq \text{vel\_fast\_thresh}, \\
\text{Fast}, & v > \text{vel\_fast\_thresh}.
\end{cases}
$$
The direction and acceleration level can be defined similarly with corresponding thresholds.
Both grippers are processed in parallel, and the resulting discrete labels are combined to form a comprehensive motion description for each frame, encompassing multiple kinematic dimensions \textit{(e.g. ``Movement: down, Velocity: slow, Acceleration: Decelerating'')}.
Notably, the pipeline is fully customizable and users can extend it with additional windows \textit{(e.g., jerk, angular velocity)} and reorder computation stages to fit task-specific needs across diverse manipulation scenarios.
By organizing dense frame-wise kinematics into interpretable motion attributes, this pipeline enriches manipulation datasets with detailed motion priors, which are essential for robust policy learning and in-depth motion behavior analysis.

\begin{table*}[t]
\centering
\caption{
  Simulation task settings.
}
\label{tab:simulation_tasks}
\setlength{\tabcolsep}{13pt}
\begin{tabular}{lcc}
\toprule
\textbf{Task} & \textbf{Trajectories} & \textbf{Example Instruction} \\
\midrule
Blocks Ranking RGB & 50 & Arrange the red block, the green block, and the blue block from left to right in a row. \\
Grab Roller & 50 & Take hold of the roller using your arms. \\
Handover Block & 50 & Pass the red block from the left to the right. \\
Handover Mic & 50 & Pick the microphone and transfer it to the other arm. \\
Hanging Mug & 50 & Use the left arm to grab the mug, rotate it, set it down, then hang the mug onto the rack. \\
Lift Pot & 50 & Hold the pot firmly, then lift. \\
Place Bread Basket & 50 & Pick up the first bread and put it in the basket. \\
Place Bread Skillet & 50 & Grab the bread and place it inside the skillet. \\
Place Burger Fries & 50 & Use both arms to move the hamburg and the frenchfries to the tray. \\
Place Can Basket & 50 & Place the can in the basket and then lift the basket. \\
Place Can Plasticbox & 50 & Use both arms to move the left can and right can into the plasticbox. \\
Place Dual Shoes & 50 & Put two shoes, tips to the left, into the shoebox. \\
Put Object Cabinet & 50 & Open the cabinet's drawer and place the object inside it. \\
Stack Blocks Two & 50 & Shift the red block and the green block to the center and stack the green block on the red block. \\
Stack Bowls Two & 50 & Lift the first bowl, put it on the table, and stack the second bowl above. \\
\bottomrule
\end{tabular}
\end{table*}

\begin{table}[t]
\centering
\caption{
  Real-world task settings.
}
\label{tab:real_world_tasks}
\setlength{\tabcolsep}{2.5pt}
\begin{tabular}{lcc}
\toprule
\textbf{Task} & \textbf{Trajectories} & \textbf{Example Instruction} \\
\midrule
Block Basket & 50 & Put the block into the basket.  \\
Peach Drawer & 50 & Put the peach in the drawer and close it. \\
Towel Basket & 50 & Put the towel into the basket. \\
Pass Bowl & 50 & Pass the bowl from the left to the right. \\
Peach Box & 50 & Open the box and put the peach in it. \\
Fold Towel & 50 & Fold the towel. \\
Grape Plate & 298 & Pick the grapes and place them on the plate. \\
Bread Bowl & 997 & Pick the bread and put it in the bowl. \\
\bottomrule
\end{tabular}
\end{table}

\noindent\textbf{Human Verification.}
To ensure annotation quality, we conduct human verification for both trajectory- and segment-level labels.
Manual verification confirms that trajectory-level annotations achieve an overall accuracy of \textbf{84.2\%}.
The remaining errors include missed objects \textit{(11.2\%)}, localization errors \textit{(4.4\%)}, and redundant objects \textit{(0.2\%)}.
For segment-level annotations, we use an online verification mechanism, where annotators check and adjust segment boundaries and action labels in real time to ensure consistency.
Frame-level kinematic labels require no manual verification, as they are directly computed from accurate robot end-effector poses, ensuring numerical precision.
This two-stage verification effectively yields reliable and consistent labels to support robust model training and cross-embodiment generalization in robotic manipulation.

\section{Experimental Task Setups}
\label{appendix:tasks}

To evaluate the generalization and robustness of robotic manipulation algorithms across different environments, we design a dual-environment task setup comprising simulation  and real-world tasks. 
Simulation tasks are tailored for cross embodiment transfer testing, while real-world tasks serve as the benchmark for evaluating hierarchical capability pyramid. 
The simulation environment comprises 16 distinct manipulation tasks, while the real-world environment includes 8 such tasks, each task is associated with 50 trajectories.
Critically, these tasks are constructed with varying difficulty levels and a core focus on bimanual collaboration, covering a diverse range of manipulation paradigms for systematic evaluation.

\begin{figure*}[t]
  \centering
  \includegraphics[width=0.99\textwidth]{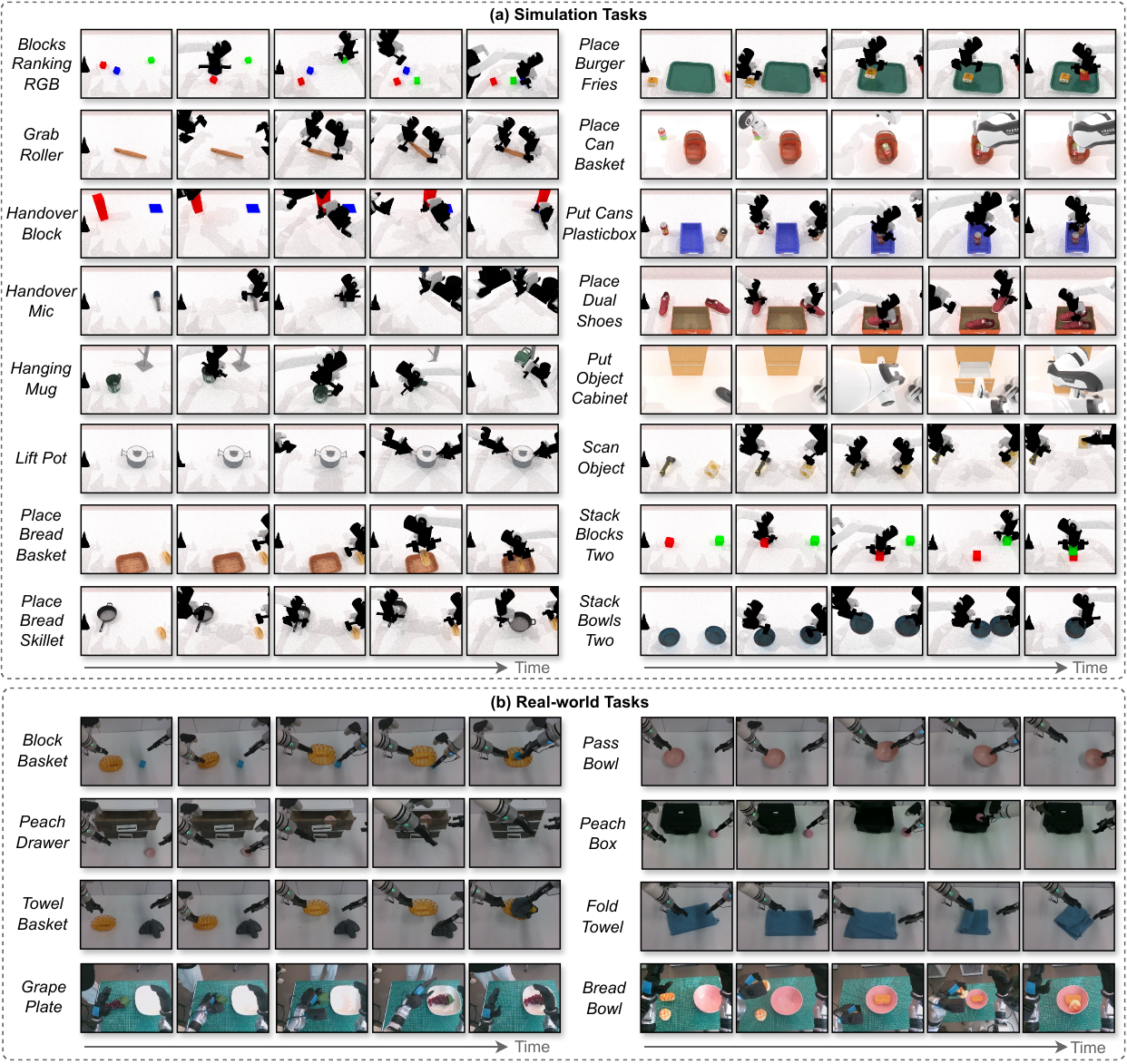}
  \caption{
    \textbf{Example task setups} for (a) simulation environment, and (b) real-world environment.
  }
  \label{fig:task_demo}
\end{figure*}
\begin{figure}[t]
  \centering
  \includegraphics[width=0.5\textwidth]{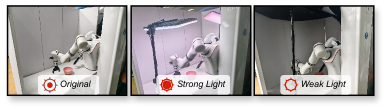}
  \caption{
    \textbf{Out-of-distribution lighting variations} in the real-world task setup, including \textit{(Left)} original lighting, \textit{(Middle)} strong lighting, and \textit{(Right)} weak lighting.
  }
  \label{fig:ood_lighting}
\end{figure}

\noindent\textbf{Simulation Task Setups.} 
As detailed in Table \ref{tab:simulation_tasks}, the simulation task suite encompasses a wide variety of manipulation scenarios.
The simulation tasks are implemented within the \textit{RoboTwin 2.0} \cite{chen2025robotwin} virtual environment, where data collection is conducted using both \textit{ARX-X5} and \textit{Franka Emika Panda} dual-arm robots, both with an inter-arm distance of 0.8 meters.
50 trajectories are collected for each task, along with corresponding state information and natural language instructions.
The simulation setup adopts a clean environment configuration without introducing background variations to ensure controlled evaluation; 
object deformation in simulation is limited, as most objects are rigid bodies, while deformable objects only feature a small number of movable joints \textit{(e.g., drawer slides)}. 
Visual sequences for the simulation environment are presented in Figure \ref{fig:task_demo} (a).

\noindent\textbf{Real-World Task Setups.}
As shown in Table \ref{tab:real_world_tasks}, the real-world task suite includes a range of manipulation scenarios with varying object properties and collaboration requirements.
The real-world tasks are executed on the \textit{Realman RMC-AIDA-L} half-humanoid robot and the \textit{Unitree G1edu-u3} humanoid robot.
A total of 50 trajectories are collected for each of the 6 tasks in the hierarchical capability pyramid experiments, while 298 and 997 trajectories are gathered for the two respective tasks in the RTML experiments, along with corresponding state information and natural language instructions.
Real-world experimental outcomes are inherently influenced by practical factors including the robot's control precision and ambient environmental noise, which introduces realistic variability and complexity absent in simulation. 
Task demonstrations for the real-world environment are shown in Figure \ref{fig:task_demo} (b).

\noindent\textbf{Out-of-Distribution Lighting Variations.}
To further enhance the robustness evaluation of manipulation algorithms under real-world visual variations, we introduce out-of-distribution (OOD) conditions during real-world task execution, and Figure \ref{fig:ood_lighting} illustrates the lighting variations applied to the real-world task setup.
This figure depicts three distinct lighting scenarios: \textit{(Left) the original lighting, (Middle) strong lighting, and (Right) weak lighting}. 
These lighting variations introduce significant visual challenges, including uneven illumination, cast shadows, reduced object texture visibility, and brightness disparities, which directly impact the performance of vision-based perception modules \textit{(e.g., object detection, pose estimation)} critical for robotic manipulation. 
Notably, there are other OOD conditions in our real-world setup, such as object color changes and background variations.
By incorporating such environment diversity into the real-world task setup, we ensure that the dataset and evaluation framework rigorously test the algorithm's ability to generalize across non-ideal conditions, which is essential for real-world deployment.

\begin{figure*}[h!]
  \centering
  \includegraphics[width=\textwidth]{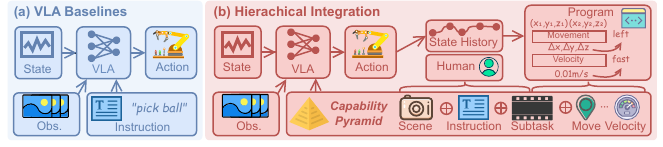}
  \caption{
    \textbf{Hierarchical Capability Pyramid Integration.}
    (a) Vision-Language-Action (VLA) Baseline Model Implementation.
    The model directly maps multi-modal inputs (state, observation, instruction) to primitive actions without explicit hierarchical knowledge integration.
    (b) Hierarchical Integration Implementation.
    The framework incorporates hierarchical annotations into a structured decision-making pipeline, enhancing action generation with multi-level task knowledge and motion constraints.
  }
  \label{fig:hai_implementation}
\end{figure*}

\section{Hierarchical Capability Pyramid Integration}
\label{appendix:hai}

Figure \ref{fig:hai_implementation} illustrates the implementation framework of the proposed hierarchical annotation integration approach, alongside a baseline Vision-Language-Action (VLA) model for comparative analysis.

\noindent\textbf{Vision-Language-Action Baseline Model.}
As depicted in \ref{fig:hai_implementation} (a), the VLA baseline adheres to a standard end-to-end paradigm:
it ingests raw state information, visual observations, and natural language instructions as multi-modal inputs, directly maps these modalities to robotic control signals via the VLA model, and outputs primitive manipulation actions.
This approach relies solely on the model's learned representations to infer appropriate actions, without leveraging any explicit hierarchical task knowledge or motion constraints.

\noindent\textbf{Hierarchical Integration.}
Figure \ref{fig:hai_implementation} (b) presents the proposed framework, which augments the VLA pipeline by embedding hierarchical annotations into a structured decision-making pipeline.
Our core objective is to validate the effectiveness of the hierarchical capability pyramid with a minimal, non-intrusive approach that can be seamlessly integrated into existing VLA models without altering their internal architecture or learned parameters.
During inference, hierarchical task-action descriptions are constructed by fusing human instructions and an automated program module, with information derived at three complementary levels:
\begin{itemize}
    \item \textbf{Trajectory-level information.}
    The overall trajectory structure is provided by human demonstrations, which defines the complete execution flow and high-level goal of the manipulation task.

    \item \textbf{Segment-level information.}
    Atomic subtask descriptions are predefined manually, consistent with the segment-level annotation pipeline introduced earlier.
    During inference, a rule-based program automatically identifies critical switching frames based on gripper states and task progress, enabling stable and smooth transitions between consecutive subtasks.
    More details on the switching point identification can be found in the
    \textbf{Segment-Level Annotation Pipeline in Appendix \ref{appendix:annotation}}.

    \item \textbf{Frame-level information.}
    From historical pose sequences, the cascaded sliding-window scheme extracts instantaneous kinematics including displacement, velocity and acceleration.
    These continuous metrics are further converted into concise natural language motion descriptions via predefined thresholds, providing explicit, frame-wise constraints on movement direction, speed and acceleration trend.
    More details on the frame-level annotation process can be found in the
    \textbf{Frame-Level Annotation Pipeline in Appendix \ref{appendix:annotation}}.
\end{itemize}
By integrating such multi-level hierarchical annotations, the framework injects structured task priors and fine-grained motion constraints into the input space of the VLA model, enabling it to generate actions that are both semantically consistent and kinematically feasible.
We emphasize that this lightweight integration strategy serves primarily as a naive and proof-of-concept to verify the value of hierarchical priors, and we believe more effective integration methods can be explored in future work to further enhance the synergy between hierarchical knowledge and VLA models.

\onecolumn
\begin{lstlisting}[style=yaml, caption={RTML Example "pull bowl storage bread"}, label={lst:rtml_example}]
# RTML V1.0
task:
  id: "pull_bowl_storage_bread"

  # Global constraints
  global_constraints:
    velocity:
      linear:
        max: 0.5          # m/s 
        mean_max: 0.3     # m/s 
    acceleration:
      linear:
        max: 12.0          # m/s2

  # Local stage constraints
  stages:
    - id: "move_bowl_right"
      match_subtask: "Move the pink bowl to the center of table with right hand"
      constraints:
        workspace:
          right:
            min: [0.10, -0.40, 0.10]
            max: [0.25, -0.20, 0.30]
        velocity:
          linear:
            mean_max: 0.10
            std_max: 0.08
        idle_arm:
          arm: "left"
          velocity_linear_mean_max: 0.05
        temporal:
          duration_min: 2.0
          duration_max: 6.0

    - id: "grasp_long_bread_left"
      match_subtask: "Grasp the long bread with left hand"
      constraints:
        workspace:
          left:
            min: [0.05, -0.05, -0.05]
            max: [0.25, 0.35, 0.20]
        orientation:
          left:
            angular_mean_deviation_max: 0.8
            std_max: [0.5, 0.5, 0.8]
            angular_variance_max: 0.15
        velocity:
          linear:
            mean_max: 0.12
            std_max: 0.10
        idle_arm:
          arm: "right"
          velocity_linear_mean_max: 0.05
        temporal:
          duration_min: 2.0
          duration_max: 8.0

    - id: "place_long_bread_in_bowl"
      match_subtask: "Place the long bread in pink bowl with left hand"
      constraints:
        workspace:
          left:
            min: [0.05, -0.05, -0.05]
            max: [0.25, 0.35, 0.20]
        velocity:
          linear:
            mean_max: 0.15
            std_max: 0.15
        idle_arm:
          arm: "right"
          velocity_linear_mean_max: 0.05
        temporal:
          duration_min: 1.0
          duration_max: 4.0

    - id: "grasp_round_bread_left"
      match_subtask: "Grasp the round bread with left hand"
      constraints:
        workspace:
          left:
            min: [0.05, 0.00, -0.05]
            max: [0.25, 0.35, 0.20]
        orientation:
          left:
            angular_mean_deviation_max: 0.5
            std_max: [0.5, 0.5, 0.5]
            angular_variance_max: 0.15
        velocity:
          linear:
            mean_max: 0.12
            std_max: 0.10
        idle_arm:
          arm: "right"
          velocity_linear_mean_max: 0.05
        temporal:
          duration_min: 2.0
          duration_max: 8.0

    - id: "place_round_bread_in_bowl"
      match_subtask: "Place the round bread in pink bowl with left hand"
      constraints:
        workspace:
          left:
            min: [0.05, 0.00, -0.05]
            max: [0.25, 0.35, 0.20]
        velocity:
          linear:
            mean_max: 0.15
            std_max: 0.15
        idle_arm:
          arm: "right"
          velocity_linear_mean_max: 0.05
        temporal:
          duration_min: 1.0
          duration_max: 4.0

    - id: "End"
      match_subtask: "End"
      constraints:
        velocity:
          linear:
            mean_max: 0.12
            std_max: 0.12
        temporal:
          duration_max: 6.0
\end{lstlisting}

\end{document}